\pdfoutput=1
\documentclass[11pt]{article}

\usepackage[]{EMNLP2023}

\usepackage{times}
\usepackage{latexsym}

\usepackage[T1]{fontenc}
\usepackage[utf8]{inputenc}

\usepackage{microtype}

\usepackage{inconsolata}

\usepackage{multirow}
\usepackage{booktabs}
\usepackage{color}
\usepackage{makecell}
\usepackage{hyperref}
\usepackage{graphicx}
\usepackage {amsmath}
\usepackage{amsfonts,amssymb} 
\usepackage{algorithm}
\usepackage{algorithmic}
\usepackage{float}
\newcommand{\ours}{IdealGPT}
\newcommand{\bfours}{\textbf{IdealGPT}}
\newcommand*\samethanks[1][\value{footnote}]{\footnotemark[#1]}
\newcommand{\ie}{\textit{i}.\textit{e}.}
\newcommand{\eg}{\textit{e}.\textit{g}.}
\definecolor{lb}{gray}{0.6}
\newcommand{\lb}[1]{\textcolor{lb}{ #1}}

\title{\ours: Iteratively Decomposing Vision and Language Reasoning \\ via Large Language Models}
\author{Haoxuan You\textsuperscript{1}\thanks{\indent Equal Contribution}, Rui Sun\textsuperscript{1}\samethanks[1], Zhecan Wang\textsuperscript{1}\samethanks[1], Long Chen\textsuperscript{2}, Gengyu Wang\textsuperscript{1}
\\
\textbf{Hammad A. Ayyubi\textsuperscript{1}, 
Kai-Wei Chang\textsuperscript{3}, Shih-Fu Chang\textsuperscript{1}}
\\
\small \textsuperscript{1} Columbia University \quad
\small \textsuperscript{2} HKUST \quad
\small \textsuperscript{3} University of California, Los Angeles \\
\small \texttt{\{hy2612, rs4110, zw2627, gengyu.wang, ha2578, sc250\}@columbia.edu}\\
\small \texttt{longchen@ust.hk, kwchang@cs.ucla.edu} 
}

\begin{document}
\maketitle
\begin{abstract}

% While some efforts have been made in decompositional reasoning process, existing sub-question generation methods are not generalizable and lack commonsense reasoning ability.

The field of vision-and-language (VL) understanding has made unprecedented progress with end-to-end large pre-trained VL models (VLMs). However, they still fall short in zero-shot reasoning tasks that require \emph{multi-step} inferencing. To achieve this goal, previous works resort to a divide-and-conquer pipeline. In this paper, we argue that previous efforts have several inherent shortcomings: 1) They rely on domain-specific sub-question decomposing models. 2) They force models to predict the final answer even if the sub-questions or sub-answers provide insufficient information. 
% To this end, 
We address these limitations via \bfours{}, a framework that \emph{iteratively} decomposes VL reasoning using large language models (LLMs). Specifically, \ours{} utilizes an LLM to generate sub-questions, a VLM to provide corresponding sub-answers, and another LLM to reason to achieve the final answer. These three modules perform the divide-and-conquer procedure iteratively until the model is confident about the final answer to the main question. We evaluate \ours{} on multiple challenging VL reasoning tasks under a zero-shot setting. In particular, our \ours{} outperforms the best existing GPT-4-like models by an absolute 10\% on VCR and 15\% on SNLI-VE.  Code is available at \href{https://github.com/Hxyou/IdealGPT}{https://github.com/Hxyou/IdealGPT}.

\end{abstract}

\section{Introduction}

\begin{figure}[t]
        \includegraphics[width=0.32\textheight]{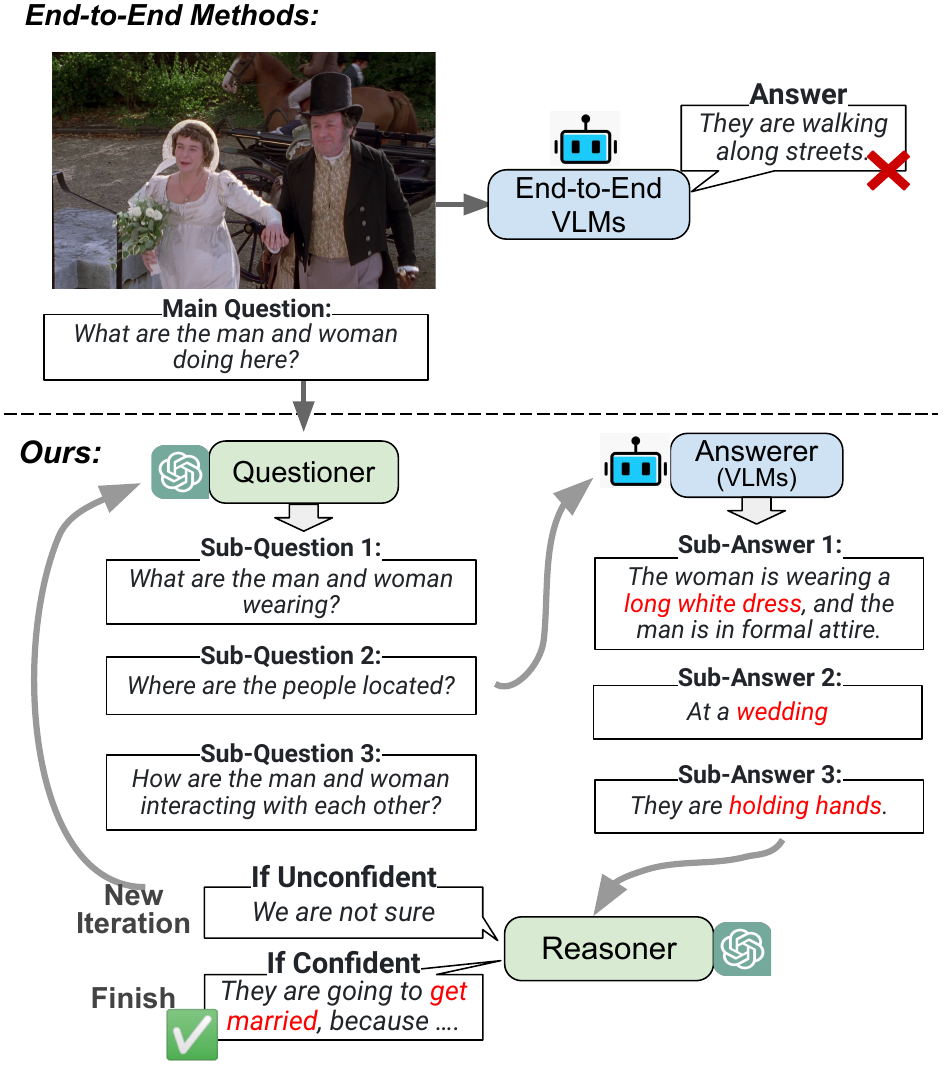}
        \vspace{-5mm}
        \captionof{figure}{Comparisons between the pipelines of prevalent end-to-end VLMs (Upper) and our proposed \ours{} (Below) for VL reasoning tasks.}
\label{fig:1}
\end{figure}

The field of vision-and-language (VL) understanding has witnessed a proliferation of pre-trained VL models (VLMs)~\citep{yu2022coca, you2023cobit, alayrac2022flamingo, zhu2023minigpt, liu2023visual}. They are usually pre-trained and fine-tuned in an end-to-end fashion, \ie, these models always directly make final predictions in a single step. With abundant pre-trained knowledge, they already achieve impressive results in comparison to human performance across many downstream VL tasks. However, they still struggle to address zero-shot VL reasoning tasks that require intricate or multi-step inferencing such as visual commonsense reasoning (VCR)~\cite{zellers2019recognition}, as exemplified in Figure~\ref{fig:1}. Despite the overall success of these models, the difficulties with zero-shot reasoning settings represent a serious challenge in the current state of VL research.
% \gy{feel its better to stress more on the zero-shot part, feel free to comment out or merge: 
% I like this below paragraph too -- Hammad
% However, a significant stumbling block lies in their struggle to handle zero-shot VL reasoning tasks effectively. These tasks demand intricate or multi-step inferencing, as exemplified by visual commonsense reasoning (VCR) (Zellers et al., 2019), shown in Figure 1. Despite the overall success of these models, the difficulties with zero-shot reasoning settings represent a serious and unresolved challenge in the current state of VL research.}

% However, we humans handle the challenging reasoning problem step by step in a divide-and-conquer way. For the example in Fig \ref{fig:1}, to answer the question \textit{what are the man and the woman doing here}, we first subconsciously look up several question-oriented details in the image. For instance, we probably wonder (1) what are they dressing? (2) what is the interaction between them? (3) where are they located?  After finding out the above details, we use commonsense knowledge to obtain a conclusion that \textit{they are going to get married}.  The above procedure can be explicitly formulated as a decompositional reasoning process including the following steps: divide the main question into multiple sub-questions focusing on visual details, then answer sub-questions, and finally reason upon the sub-answers to answer the main question.   

In comparison, humans are able to excel in these tasks effortlessly. As in Figure~\ref{fig:1}, to answer the question, \textit{``What are the man and woman doing here?''}, we humans would intuitively approach it in a divide-and-conquer fashion by starting with simple questions like ``What are they dressed as?'' to identify people in the image first. Then we may advance further to wonder about their interactions, ``What are they doing?'' and the location, ``Where are they at?''. Even though it may seem challenging to answer the original question directly, it is much easier to answer the three sub-questions. After answering the three simpler sub-questions, we can understand the image comprehensively, and utilize commonsense knowledge to obtain a conclusion: ``\textit{they are going to get married}''. This step-by-step procedure can be explicitly formulated as a decompositional reasoning process of three key steps: (1) dividing the main question into multiple sub-questions focusing on visual details, (2) answering easier sub-questions, and (3) reasoning upon the sub-answers to address the main question.   

Inspired by this divide-and-conquer manner, several works solve VL reasoning with this compositional pipeline: collecting the sub-questions dataset \citep{selvaraju2020squinting}, generating sub-questions for the main question \citep{uehara2022learning, wang2022co, wang2022understanding}, and utilizing sub-questions to help to answer the main question \citep{selvaraju2020squinting, uehara2022learning, wang2022co}. Nonetheless, several drawbacks still exist and hinder their practice: 1) Existing methods rely on task-specific trained models to generate sub-questions, which are not generalizable.
% (2) lack commonsense reasoning ability to analyze the obtained sub-answers to find the final answer. Moreover, since the sub-questions and sub-answers are generated by models, it happens that current sub-answers might be in conflict, and the sub-questions/sub-answers might not be informative enough to find a sure answer to the main question, which necessitates additional and supplementary sub-questions.
% \gy{Feel this point could be better explained, which kind of domain in ''domain-specific``? dataset/task specific? and what kind of models? if it's LLM models, aren't most LLMs are general models?}
% \ha{We can mention this -- ViperGPT limited by the set of available APIs, VisProg limited by the set of available commands, Neuro-symbolic VQA limited by the set of primitive operations in the Domain Specific Langauage (DSL).}
2) Existing methods impractically assume that the sub-questions and sub-answers generated in one round can guarantee sufficient evidence to address the main question, which is profoundly incorrect in practice. For instance, the generated sub-questions might be not informative enough or deviate from  the main question. These predicted sub-answers may be noisy and raise conflicts due to possible misprediction. 
% \gy{Feel having a specific example here will much better explain the point} \ha{Or cite some work which have investigated this inconsistency}
% Despite these common issues, existing methods all force the models to predict the main question given current sub-questions and sub-answers, even when the provided information is insufficient or misleading. 
Therefore, existing methods may cause irrational final predictions or be forced to learn spurious bias to guess the final answer.

To address these above-mentioned issues, we proposed a new framework \bfours{}, which \textbf{I}teratively \textbf{de}composes vision \textbf{a}nd \textbf{l}anguage reasoning with large language models (LLMs).  Specifically, \ours{} employs two LLMs (\eg, GPT \citep{ouyang2022training,openai2023gpt} in our experiments) as the \texttt{Questioner} and the \texttt{Reasoner}, and a pretrained VLM as the \texttt{Answerer}. In this framework, these three agents interact with each other to perform the divide-and-conquer procedure iteratively until finding a confident answer to the main question. As shown in Figure~\ref{fig:1}, the \texttt{Questioner} first raises sub-questions decomposed from the main question, and the \texttt{Answerer} subsequently replies with the corresponding sub-answers. Subsequently, the \texttt{Reasoner} analyzes cumulative information extracted from sub-answers to infer the possible answer to the main question. 
% Sometimes \texttt{Reasoner} does not find sufficient evidence to address the main question confidently. This could be due to uninformative sub-questions or noisy sub-answers. In either case, instead of forcing \texttt{Reasoner} to predict a final answer without confidence, the explanation/analysis from \texttt{Reasoner} would be fed into \texttt{Questioner}. 
If the \texttt{Reasoner} ascertains that the evidence gathered so far is insufficient to confidently answer the main question (either due to uninformative sub-questions or noisy sub-answers), it loops back to the \texttt{Questioner} with its analysis of the gathered evidence.
Hence, \texttt{Questioner} would purposely 
% continue 
try to generate more targeted sub-questions to obtain more informative evidence. 
These iterations of the QA loop would continue to be initiated until \texttt{Reasoner} is confident of resolving the main question or the number of iterations reaches a predefined threshold.

Compared with previous compositional and end-to-end methods, the proposed \ours{} brings several significant benefits: (1) \textbf{Transparency and Interpretability}. It is straightforward to pinpoint which sub-answer or reasoning step results in the undesired final answer. Additionally, multiple-round interactions allow models to showcase their understanding and reasoning process step by step which leads to the final answer. (2) \textbf{Modularity}. With the rapid development of LLMs and VLMs, the \texttt{Questioner}/\texttt{Reasoner} and \texttt{Answerer} can easily be updated to the more powerful LLM and VLM to improve performance. (3) \textbf{Robustness}. Existing models still heavily suffer from problems like superficial biases like syntactic bias, inconsistent predictions, or hallucination. All of these can lead to conflicted and noisy evidence during reasoning steps. Our multiple rounds can robustly consider models' both noisy and accurate predictions to merge to the most confident answer. (4) \textbf{Generalizability}. \ours{} can be seamlessly applied to multiple tasks. %Gengyu Various VL tasks require reasoning and can be inherently formatted as question-answer tasks.
This is because of the fact that various VL tasks require reasoning skills and can be inherently formatted as question-answer tasks. Moreover, \ours{} illustrates strong zero-shot ability as no training or finetuning on specific tasks is needed.

% \ha{I believe it's not very sound to claim that we bring (1) Transparency, (2) Modularity and (3) Generalizability, to the table when compared to prior compositional models (ViperGPT does all of this). I believe what we can claim is that (1) We are not limited by the set of APIs/set of operations/or DSL, (2) Robustness and (3) We can achieve higher order reasoning from low level visual-spatial understanding. We could mention in the beginning that our compositional pipeline brings all the traditional advantages like (1) Transparency, (2) Modularity and (3) Generalizability. Or we could mention at the end of paragraph. That of course, in addition to these advantages, we also reap the traditional benefit of using a compositional pipeline -- ...}
% hx: I agree, but we don't have enough time, probably will rewrite it in emnlp submission version.
We quantitatively evaluate \ours{} on several challenging VL reasoning tasks in a zero-shot setting, including VCR and Visual Entailment (SNLI-VE) \citep{xie2019visual}. Since zero-shot VCR and SNLI-VE are too challenging for most of the previous VLMs \citep{li2023blip, yu2022coca, alayrac2022flamingo}, we found only GPT-4-like models based on instruction-tuned LLMs, such as MiniGPT4 \citep{zhu2023minigpt} and LLaVA \citep{liu2023visual}, are capable of tackling the tasks. Compared with the above-mentioned GPT-4-like models, our \ours{} can outperform the best by an absolute 10\% in VCR and 15\% in SNLI-VE. 
% We further quantitatively demonstrate the generalizability of \ours{} in the task of In-the-wild VQA, where the questions/queries are daily-life instructions. 

\section{Related Works}

\subsection{Compositional QA in Vision/Language}
% \subsection{Divided-and-Conquer Reasoning in Vision/Language}
Answering multi-hop reasoning questions directly can be challenging in NLP. \citet{press2022measuring} investigates the ability of language models to perform compositional reasoning tasks where the final solution depends on correctly composing the answers. \citet{wang2022self} exploits a self-consistency method to sample a diverse set of reasoning paths and then filter and aggregate by choosing the most consistent answer. \citet{yoran2023answering} samples multiple reasoning chains and mixes information between them to select the most relevant facts in generating an explanation and predicting the answer. In VQA, in order to investigate the reasoning process and promote the reliability of models, SQuINT \citep{selvaraju2020squinting} collects VQA-introspect dataset providing low-level perception sub-questions to answer the complex reasoning questions.  Some methods \citep{uehara2022learning, wang2022co} decompose the original complicated questions into several informative sub-questions. By answering these sub-questions, they can help to answer the original questions. These existing methods rely on task-specific trained models and generate sub-questions and sub-answers in one round, which prevents their generalizability and reasoning ability in practical problems. However, \ours{} can be seamlessly utilized in different tasks by slightly adjusting the prompt. Moreover, our iterative approach can efficiently solve challenging tasks such as VCR and SNLI-VE without further training.
% \gy{could be better to have short summary of these work and a comparison with our work here}
% VQA:\citep{selvaraju2020squinting, uehara2022learning, wang2022co}
% QA (NLP): \citep{press2022measuring, wang2022self, yoran2023answering}
% Other approaches leverage Chain-of-Thought (CoT) to deal with reasoning questions.

\subsection{End-to-End Vision-Language Models}
% Without Instruction-Tuning: 
Vision-language pre-training models \citep{li2019visualbert, chen2020uniter, li2022blip, you2022learning, yu2022coca} are pre-trained on large-scale image-text pairs, which enable these models' joint understanding between different modalities. 
% Then, they are finetuned to multimodal downstream tasks.
Recently, there is a trend to utilize the knowledge from LLMs and align visual features to the text space. Flamingo \citep{alayrac2022flamingo} inserts cross-attention layers into LLMs to import visual features and employs billions of image-text pairs to pre-train the new layers. BLIP-2 \citep{li2023blip} is powered by the pre-trained visual encoder and LLMs. It uses a lightweight Querying Transformer (Q-Former) following a two-stage pre-training to bridge the modality gap. Inspired by InstructGPT \citep{ouyang2022training}, to improve the generalization performance of LLMs to unseen tasks and align users' intentions, some vision-language models finetune the pre-trained models using extra instruction-tuning data. LLaVA \citep{liu2023visual} leverages the trainable projection layer to project the output from the visual encoder to the LLM and utilizes vision-language conversational data generated by GPT-4 \citep{openai2023gpt} to finetune the LLM. MiniGPT4 \citep{zhu2023minigpt} employs the pre-trained visual encoder and Q-Former from BLIP-2 and uses image captions generated by ChatGPT to perform training on the LLM and the single linear projection layer.% InstructBLIP \citep{dai2023instructblip} gathers instruction tuning format data from publicly available datasets and instruction-tunes the BLIP-2 model with the language modeling loss to generate the response.
\subsection{GPT-Aided Visual Reasoning}

LLMs are pre-trained on colossal corpus so they can attain rich prior knowledge and strong reasoning ability. 
% However, they can only deal with unimodal tasks. 
Recently, a trend has emerged that leverages LLMs in combination with a range of vision or multimodal models. These approaches create a system capable of addressing various multimodal tasks without the need for additional training. MM-React \citep{yang2023mm}, HuggingGPT \citep{shen2023hugginggpt}, Chameleon \citep{lu2023chameleon}, Visual ChatGPT \citep{wu2023visual} regard GPT (i.e., ChatGPT, GPT-4) as a controller to coordinate and collaborate with other models (e.g., visual foundation models) to tackle complicated multimodal tasks. VisProg \citep{gupta2022visual} and ViperGPT \citep{suris2023vipergpt} exploit GPT-3 \citep{brown2020language} and CodeX \citep{chen2021evaluating} to generate a program to solve vision-language tasks in a one-round query answering. ChatCaptioner \citep{zhu2023chatgpt} lets ChatGPT and BLIP-2 interact to accomplish image captioning in a dialogue approach. All of them borrow the strong reasoning ability from LLMs and boost performance in a wide range of vision-language tasks. Different from ViperGPT, VisProg, and ChatCaptioner, \ours{} solves vision-language tasks iteratively in a multi-round manner and our proposed method can be conveniently deployed in a diverse set of vision-language tasks such as VCR and SNLI-VE. A more detailed comparison is in Sec. \ref{sec:comparison}.

% \citep{gupta2022visual, suris2023vipergpt, wu2023visual, shen2023hugginggpt, yang2023mm, lu2023chameleon, zhu2023chatgpt}

\begin{figure*}[t]
        \centering
        \includegraphics[width=0.65\textheight]{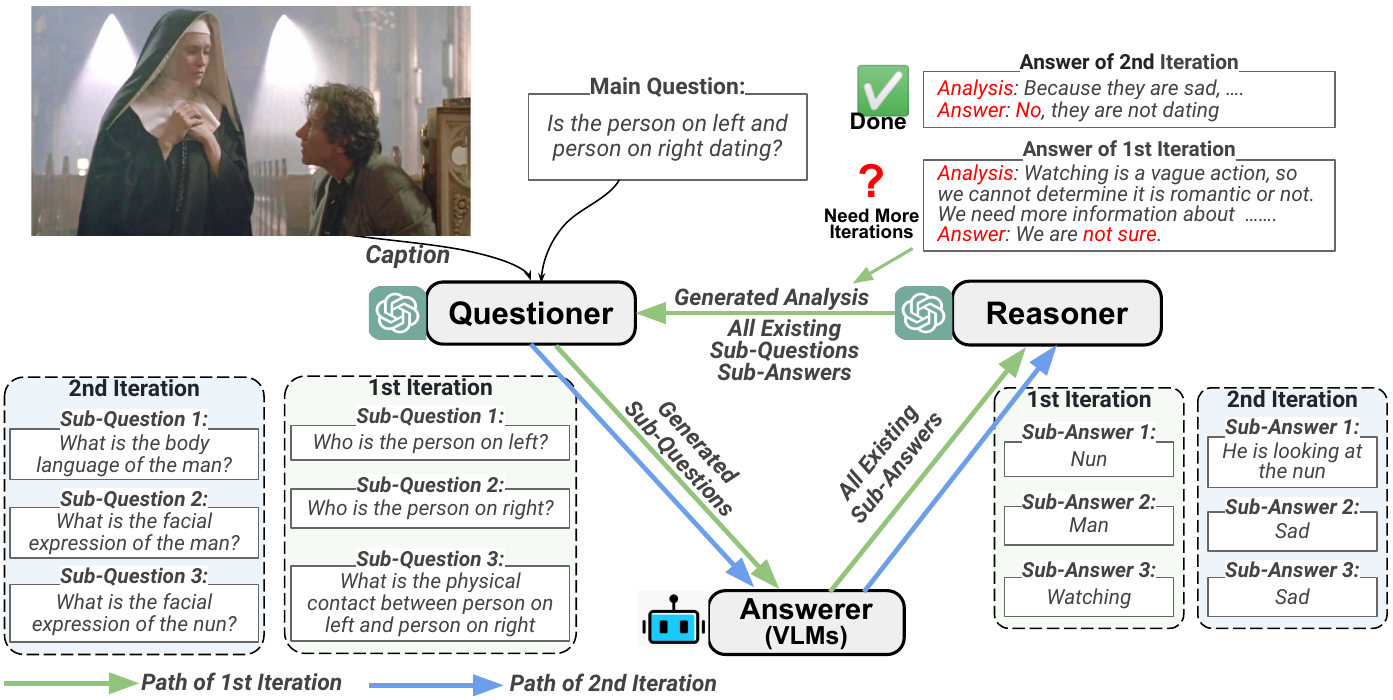}
        \caption{The pipeline of proposed \ours{}. We use an example in VCR validation set for illustration, which is finished in 2 iterations.}
\label{fig:2}
\end{figure*}

\section{Method}
In this section, we introduce the proposed \ours{}. Our focus is on the tasks of open-domain VQA, where a question $q$ is asked about an image $I$.  There are three components in \ours{} framework: a \texttt{Questioner}, an \texttt{Answerer}, and a \texttt{Reasoner}. In each iteration, based on $q$ and $I$, \texttt{Questioner} first raises a set of sub-questions $SubQ=\{sq_1, ..., sq_i\}$ (Sec. \ref{sec:questioner}), then \texttt{Answerer} generates the corresponding sub-answers $SubA=\{sa_1, ..., sa_i\}$ (Sec. \ref{sec:answerer}), and \texttt{Reasoner} analyzes both $SubA$ and $SubQ$ to decide if  a confident answer $a$ to the main question $q$ can be derived (Sec. \ref{sec:reasoner}). If a confident answer cannot be inferred in the current iteration, \texttt{Questioner} is prompted to ask additional supplementary sub-questions, and another ``\texttt{Questioner-Answerer-Reasoner}'' iteration is triggered. The above loop keeps iterating until the \texttt{Reasoner} finds a confident final answer or the number of iterations reaches a predefined maximum. The overall pipeline is shown in Figure~\ref{fig:2}.

\subsection{\texttt{Questioner}}
\label{sec:questioner}
In previous works %along this direction
\citep{uehara2022learning, wang2022co}, sub-questions are generated by models trained on specific sub-questions data \citep{selvaraju2020squinting}. However, since the training data are specifically annotated for the samples in \citet{antol2015vqa}, these sub-question generators cannot scale and generalize to different domains and complex reasoning tasks. Annotating a sub-question dataset covering all types of visual reasoning tasks and image domains is also infeasible. Recently, LLMs \citep{ouyang2022training, openai2023gpt, anil2023palm} have demonstrated a strong ability to follow instructions and reason with human knowledge. Additionally, some works \citep{suris2023vipergpt, wu2023visual, zhu2023chatgpt} have utilized LLMs to aid visual tasks, which demonstrates that LLMs have acquired diverse visual knowledge to a certain degree as well. 
% \gy{it seems too long to introduce other work in ours method section, should be simplified in one/two sentence or move to related work}
% haoxuan: I think that's fine, it can emphasize our motivation.
Inspired by these findings, we prompt GPT as a \texttt{Questioner} to raise sub-questions. By default, ChatGPT \citep{ouyang2022training} is used. While GPT-4 \citep{openai2023gpt} is a stronger alternative, it is costlier. 

The input in VQA tasks usually includes a main question $q$ and an image $I$, and sometimes answer candidates $A=\{a_1, ..., a_n\}$ if the task is formatted as a multiple-choice QA problem. The target of the \texttt{Questioner} is first to recognize the evidence needed to address $q$ and then decompose it into sub-evidences. To acquire those sub-evidences, \texttt{Questioner} would then generate a set of sub-questions $SubQ=\{sq_1, ..., sq_i\}$. For achieving quality results, we also design a prompt $P_{q}$ as an instruction for GPT to understand the objective and the desired output. For each task, the prompt is slightly different to accommodate the task descriptions\footnote{See Prompts in Appendix \ref{app:prompts}}. With solely the main question $q$ and prompt $P_{q}$ input into the \texttt{Questioner}, we empirically found that the generated sub-questions from initial iterations tend to be too generic and uninformative. 
% This may be because LLMs cannot see images such that it possesses no vision knowledge at all at the beginning. 
This could be because LLMs don't see the images and as such are devoid of the visual input.
To facilitate the \texttt{Questioner} to understand the image and generate more informative questions, we provide a short caption $C_I$ generated by a VLM  as an additional input to the \texttt{Questioner}. Therefore, in the first iteration, the sub-question generation can be formulated as follows:
\begin{equation}
\centering
SubQ_1 = \text{ChatGPT}(q, C_I, P_q).
\end{equation}

As mentioned before, there may not be sufficient evidence for the \texttt{Reasoner} to address the main question after only one iteration. 
% \ha{repetitive} 
This can be due to common issues like the sub-questions are not informative enough or conflict/noise existing among sub-answers.
% since the Answerer cannot perfectly answer every sub-questions and the Questioner cannot perfectly ask informative sub-questions every time, it sometimes happens that the Reasoner cannot infer a confident answer to the main question. 
In this case, \ours{} would prompt the \texttt{Reasoner} to generate an explanation/analysis regarding why it may not have sufficient evidence to address the main question. Subsequently, we would loop back to the \texttt{Questioner} to generate additional supplementary sub-questions. In the $t$-th iteration ($t>1$), \texttt{Questioner} accepts all previous sub-questions $SubQ_{1:t-1}$ and sub-answers $SubA_{1:t-1}$, and the previous analysis $E_{t-1}$ from \texttt{Reasoner}\footnote{See details in Sec. \ref{sec:reasoner}} as additional input:
\begin{equation}
\centering
\begin{aligned}
SubQ_t = & \text{ChatGPT}(q, C_I, P_q, \\
& SubQ_{1:t-1}, SubA_{1:t-1}, E_{t-1}),
\end{aligned}
\end{equation}
where $SubQ_{1:t-1} = \{SubQ_1\cup...\cup SubQ_{t-1}\}$ and $SubA_{1:t-1} = \{SubA_1\cup...\cup SubA_{t-1}\}$. Previous sub-questions and sub-answers can inform \texttt{Questioner} what has been asked and solved, and the analysis can guide \texttt{Questioner} to generate more specific sub-questions, such as sub-questions to collect more informative evidence about a specific visual object and so on.

\subsection{\texttt{Answerer}}
\label{sec:answerer}
Given the generated sub-questions $SubQ$, the goal of \texttt{Answerer} is to answer them correspondingly to provide evidence for answering the main question. In \ours, the \texttt{Answerer} is a pre-trained VLM without finetuning on any dataset to keep intact its generalization ability. Each sub-question is answered separately:
\begin{equation}
\centering
sa_i = \text{VLM}(sq_i, I),
\end{equation}
where  $sq_i \in SubQ$ and  $sa_i \in SubA$. It is noted that theoretically, the \texttt{Answerer} can not only be end-to-end VLMs but also VL systems such as \citet{suris2023vipergpt, gupta2022visual, shen2023hugginggpt}. 
% We experiment with several VLMs including BLIP2 \citep{li2023blip}, MiniGPT4\cite{zhu2023minigpt}, LLaVA\cite{liu2023visual} and observe BLIP2 as the best choice among them in both effectiveness and efficiency.

\subsection{\texttt{Reasoner}}
\label{sec:reasoner}
GPT-like LLMs \citep{ouyang2022training, openai2023gpt, anil2023palm} have shown impressive summarization and reasoning ability with commonsense knowledge. Therefore, like \texttt{Questioner}, we also choose ChatGPT as the \texttt{Reasoner} but prompt it differently. Specifically, the input to \texttt{Reasoner} contains main question $q$, caption $C_I$,  all existing sub-questions $SubQ_{1:t}$ and corresponding sub-answers $SubA_{1:t}$. And the \texttt{Reasoner} is prompted to generate both the analysis and the final answer with its prompt $P_R$\footnote{See Prompts in Appendix \ref{app:prompts}}:
\begin{equation}
\centering
E_t, a = \text{ChatGPT}(SubQ_{1:t}, SubA_{1:t}, q, C_I, P_R).
\end{equation}
If the \texttt{Reasoner} is not confident about the final answer, it is instructed to faithfully indicate that by generating a specific response such as ``We are not sure''. If this particular response is detected, we start another iteration by asking the \texttt{Questioner} to add supplementary sub-questions. The above procedure forms a loop among the three agents, which will stop if the \texttt{Reasoner} can find a confident answer or the number of iterations reaches a pre-defined bound (a hyperparameter). 

\subsection{Comparison with Other Methods}\label{sec:comparison}
\noindent\textbf{\textit{v.s.}ViperGPT/VisProg}. VisProg \citep{gupta2022visual} and ViperGPT \citep{suris2023vipergpt} utilize LLMs to decompose VL tasks into steps of explicit conditional logic in coding based on low-level visual-spatial detection. \ours{} shares a similar idea of decomposition or divide-and-conquer. However, \ours{} can iteratively go through the divide-and-conquer process until collecting sufficient evidence to generate a confident answer. This multi-pass process involves the continuous refinement of the set of sub-questions and even a re-correctifying mechanism to the final answer prediction. Conversely, ViperGPT/VisProg performs programs in one pass regardless of whether the predicted answer is confident or not. This difference also applies to  \citet{yang2023mm, shen2023hugginggpt,lu2023chameleon}. Moreover, ViperGPT is limited by the set of available APIs, VisProg is limited by the set of available commands, and Neuro-symbolic VQA is limited by the set of primitive operations in the Domain Specific Language (DSL).

\noindent\textbf{\textit{v.s.}ChatCaptioner}. ChatCaptioner \citep{zhu2023chatgpt} lets ChatGPT and BLIP-2 interact with each other to caption images in a dialogue system. \ours{} shares the idea of utilizing ChatGPT to generate questions and VLMs to answer them. But \ours{} focuses on the VL reasoning tasks and incorporates the iterative design for generalized zero-shot vision language reasoning.

% dealing with the confidence issue of answering. 

\begin{figure}[t]
        \includegraphics[width=0.32\textheight]{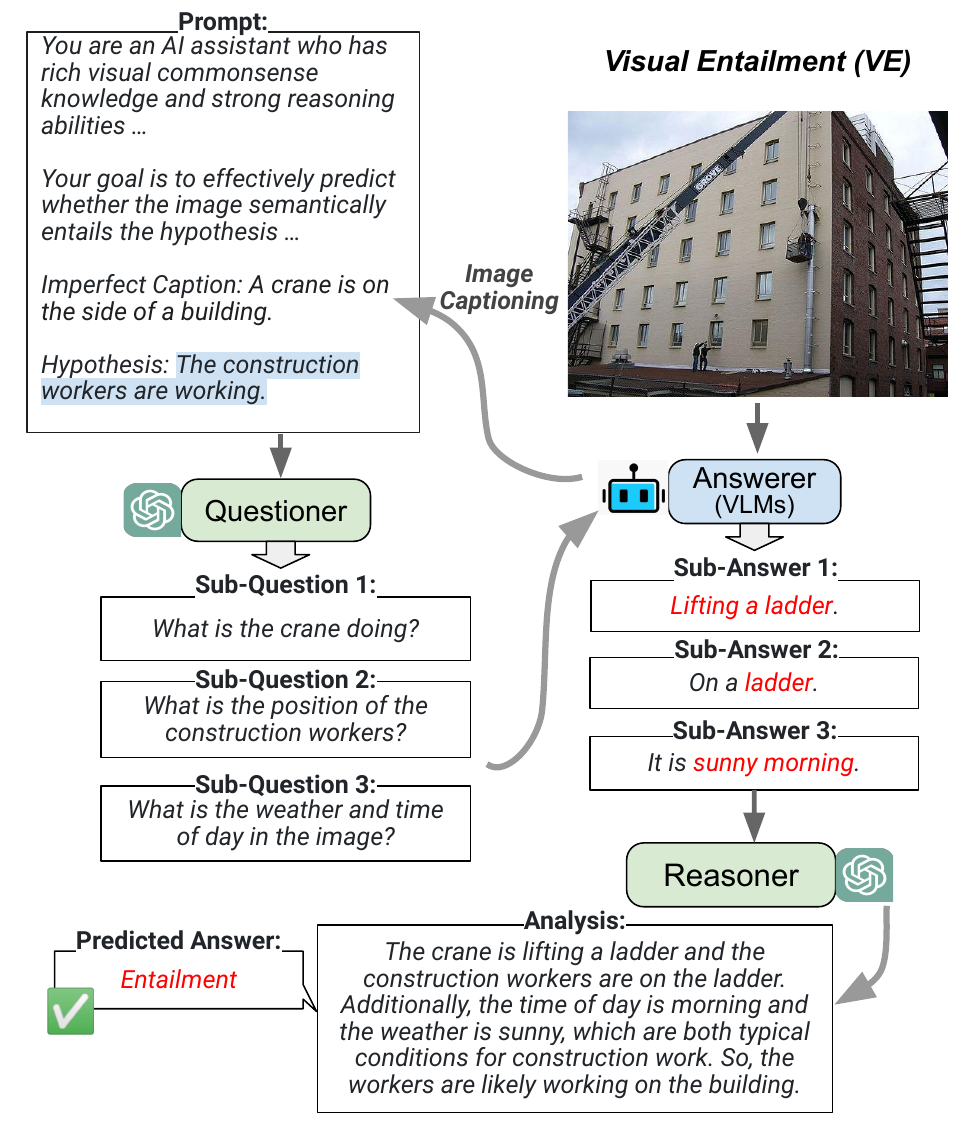}
        \vspace{-5mm}
        \captionof{figure}{The illustration of how proposed \ours{} works in SNLI-VE. \ours{} exhibits its perceptional ability in sub-question 1 and 2. Plus, it shows a strong commonsense reasoning ability in sub-question 3.}
\label{fig:3}
\end{figure}

\section{Experiments}

In this section, we evaluate our method quantitatively on two main tasks - Visual Commonsense Reasoning (VCR) and Visual Entailment (SNLI-VE). We introduce datasets and models first. Then, we show superior zero-shot performance on SNLI-VE and VCR compared to other existing models. Afterward, ablation studies are conducted to confirm the effectiveness of our proposed method.

\subsection{Experimental Setup}
\textbf{Datasets.} In this paper, we employ two VL datasets, VCR \citep{zellers2019recognition} and SNLI-VE \citep{xie2019visual} as they represent the two typical VL reasoning tasks, visual question answering and visual entailment. 
% Humans perform this task by verifying if mentioned visual facts exist in images one by one, which is essentially a decompositional fact verification process.   
Different from the traditional VQA \citep{antol2015vqa} task, VCR needs commonsense knowledge to understand the visual scenes and requires multiple-steps reasoning to answer the question. Also, in terms of the task format, it is a multiple-choice QA problem. 
SNLI-VE originates from Stanford Natural Language Inference (SNLI) \citep{bowman2015large}, a text entailment (TE) task based on Flicker30k \citep{young2014image}. It extends TE into the visual domain and asks the model whether the image is semantically entailed/neutral/contradicted to the text hypothesis, thus it can be treated as a three-category classification task.
In VCR and SNLI-VE, we randomly select 5000 samples from the val/dev split of the dataset and evaluate our method in the \textbf{zero-shot} scenario with the accuracy for evaluation.

\noindent\textbf{Models.} We choose ChatGPT to act as the Reasoner and \texttt{Questioner} and access it via ``gpt-3.5-turbo'' API. It should be noted that we set the temperature as 0 to reduce the randomness so that we can have a stable result to see how our proposed method performs in different tasks. As for the \texttt{Answerer}, three pre-trained VLMs (BLIP-2, MiniGPT4, and LLaVA) are selected to serve for a comprehensive comparison. It's noted that all VLMs we use are pre-trained-only without being finetuned on any dataset to guarantee zero-shot generalizability.  We design three simple prompts for these models respectively (more details can be found in the appendix \ref{app:prompts}) to help them better answer sub-questions. Further, in practice, these VLMs are also used to produce image captions so that the ChatGPT can reserve a general understanding of the unseen image initially.

\subsection{Visual Commonsense Reasoning}

\begin{table}[t]
\begin{center}
\renewcommand\tabcolsep{3pt}
\scalebox{0.9}{
\begin{tabular}{llc}
    \toprule
    & & Acc.(\%)\\
    \midrule
    & Random Guess & 25 \\
    \midrule
    \multirow{3}{*}{\rotatebox{90}{\lb{Sup.}}}& \lb{R2C} \citep{zellers2019recognition} & \lb{63.8} \\
    & \lb{VisualBERT} \citep{li2019visualbert} & \lb{70.8} \\
    & \lb{MerlotReserve} \citep{zellers2022merlot} & \lb{84.0} \\
    \midrule
    \multirow{4}{*}{\rotatebox{90}{ZS.}} & BLIP-2 \cite{li2023blip} & -\\ 
    & MiniGPT4 \cite{zhu2023minigpt} & 40.6 \\
    & LLaVA \cite{liu2023visual} &  28.3\\
    & \textbf{\ours{}(ours)} & \textbf{50.7}\\
    \bottomrule
\end{tabular}
}
\end{center}
\caption{Accuracy of VCR Q$\rightarrow$A task (ZS: Zero-Shot, Sup: Supervised)}
\label{tab:vcr}
\end{table}

VCR covers various commonsense in diverse visual scenes, including temporal, spatial, causal, and mental commonsense, etc. It's formatted as a multiple-choice answering problem, and to find the correct answer, the model is expected to reason among four answer candidates to find the correct answer. In VCR, it often happens that there are multiple persons in one image, and if the question mentions one of them, the bounding box will be used to distinguish it from others. However, most existing VLMs cannot understand bounding boxes in the text input, making them hard to perform VCR directly. To alleviate that, we conduct a pre-processing to describe the mentioned person's spatial coordinate in words easy-to-understand. Please see details in Appendix \ref{app:vcr_process}.

% \hx{add pre-processing steps.}

Although a lot of models can be finetuned on this task \cite{zellers2019recognition, li2019visualbert}, there is hardly any model that tackles it in the zero-shot fashion. We empirically tried BLIP2, MiniGPT4, and LLaVA and found that only the GPT4-like models can functionally perform zero-shot VCR, while other models such as BLIP2 fail to understand the long context and the instruction of finding the correct answer.  We present the experimental result in Tab. \ref{tab:vcr}, where all zero-shot results are obtained from the randomly sampled 5000 data in the validation set. As we can see, \ours{} can outperform the best GPT4-like model, MiniGPT4 by over 10\%. It's noted that in \ours{} reported here, BLIP2-FlanT5-XL is used as the \texttt{Answerer}. Please refer to Sec. \ref{sec:ablate} for ablations on the choice of \texttt{Answerer}. 

In Fig.\ref{fig:2}, we showcase an example of how \ours{} solves an example in VCR successfully. As we can see, in the first pass, the generated sub-questions and predicted sub-answers are not informative enough to support a solid conclusion because their identities and interaction of ``watching"" cannot quite indicate whether they are dating or not. Further, in the second pass, after inputting the analysis from the \texttt{Reasoner} and existing sub-questions and sub-answers, the \texttt{Questioner} is prompted to ask additional supplementary sub-questions to collect more evidence about their expressions and body language. As a result, the updated sub-answers of sad expressions and distant body language allow the \texttt{Reasoner} to ensure a confident final answer. 

% \hx{Add a case to explain our method, also emphasize the explanation ability.}

\begin{table}[t]
\begin{center}
\renewcommand\tabcolsep{3pt}
\scalebox{0.9}{
\begin{tabular}{llc}
    \toprule
    & & Acc.(\%)\\
    \midrule
    & Random Guess & 33.3 \\
    \midrule
    \multirow{3}{*}{\rotatebox{90}{\lb{Sup.}}}& \lb{EVE-Image} \citep{xie2019visual} & \lb{71.6} \\
    & \lb{UNITER} \citep{chen2020uniter} & \lb{79.4} \\
    & \lb{OFA} \citep{wang2022ofa} & \lb{91.0} \\
    \midrule
    \multirow{4}{*}{\rotatebox{90}{ZS.}} & BLIP-2 \cite{li2023blip} & -\\ 
    & MiniGPT4 \cite{zhu2023minigpt} & 35.1 \\
    & LLaVA \cite{liu2023visual} & 40.3 \\
    & \textbf{\ours{}(ours)} & \textbf{55.3} \\
    \bottomrule
\end{tabular}
}
\end{center}
\caption{Accuracy of SNLI-VE (ZS: Zero-Shot, Sup: Supervised)}
\label{tab:ve}
\end{table}

\subsection{Visual Entailment} 
Visual Entailment requires the model to predict whether the image semantically entails the text. In each sample, there is a pair of an image and a text hypothesis along with three answer candidates (\ie, entailment, neutral, and contradiction). The model needs to select one of them to represent the relationship between the image and the text hypothesis. It is challenging to solve the three-category classification task in a zero-shot manner. As shown in Fig.\ref{fig:3}, to begin with, we make some rules and introduce the goal of this task to ChatGPT\footnote{See Prompts in Appendix \ref{app:prompts}} to ensure it can understand our instructions and respond reasonably. We utilize VLMs to generate the image caption and inject it into the prompt. Thereafter, \texttt{Questioner} decomposes the original hypothesis into several sub-questions. After answering all sub-questions by VLMs, \texttt{Reasoner} will summarize the image caption, hypothesis, all sub-questions, and corresponding sub-answers together to provide a comprehensive analysis. In Fig.\ref{fig:3}, we should notice that not only does \ours{} exhibit the perceptional ability in sub-question 1 and 2, but also it shows the strong commonsense reasoning ability in sub-question 3.

Supervised methods for SNLI-VE have been well-studied \citep{xie2019visual, chen2020uniter, wang2022ofa}, while zero-shot approaches are less explored. We tried BLIP2, LLaVA, and MiniGPT4 to do zero-shot SNLI-VE. In Tab. \ref{tab:ve}, we can observe that BLIP2 fails again on SNLI-VE. Compared to MiniGPT4 and LLaVA, \ours{} consistently surpasses them by a large margin, 20\%, and 15\% respectively. This result shows that not only can our method understand long instructions but also it is able to handle different task formats (SNLI-VE and VCR have distinctively different task formats. The former is image-hypothesis pair but the latter is the question-answering format). From Tab. \ref{tab:ve}, we can also notice that the performance of MiniGPT4 is the near random-guessing level. More details and discussion about SNLI-VE can be found in Appendix \ref{app:vedetail}.

\subsection{Ablation Studies}
\label{sec:ablate}
In this section, we ablate both the design choice and component choice. We first demonstrate the necessity of iterative design in \ours{}. Then we ablate different VLMs for generating captions and performing as \texttt{Answerer}. In all ablations, we use a randomly sampled 500 data set from VCR, which in our findings is enough to distinguish different model choices. 

\begin{table}[t]
\begin{center}
\scalebox{0.9}{
\begin{tabular}{ccc}
    \toprule
    Model & Max. \#Iterations & Acc.(\%)\\
    \midrule
    \multirow{3}{*}{\ours{}} & 1 & 49.2\\
    & 2 & 53.2\\
    & 4 & 55.8\\
    \bottomrule
\end{tabular}
}
\end{center}
\caption{Ablation of iterative decomposing. Max.\#Iterations=1 means deterministic answering in one round without iterative decomposing. }
\label{tab:iter}
\end{table}

\noindent\textbf{Iterative Decomposing.} A key design in \ours{} is that if the model (\texttt{Reasoner}) is not sure about the final answer, it will trigger a new pass of QA to ask additional sub-questions and therefore provide more visual facts to the \texttt{Reasoner}. The iterative decomposing will continue until the Reasoner is confident to make a decision or the number of passes reaches a pre-defined bound, which is a hyperparameter. We evaluate \ours{} with iterative decomposing and without iterative decomposing (\ie, Max.\#Iterations=1) in Tab. \ref{tab:iter}. We can see that iterative decomposing design can boost the accuracy by as high as 6\%. It's noted that more passes also mean more inference time, and we find setting the maximum number of iterations to 4 achieves a good trade-off between efficiency and effectiveness, which is used as default in all other experiments. Under the above setting, the average number of passes across sampled data is 1.8.  

\begin{table}[t]
\begin{center}
\scalebox{0.9}{
\begin{tabular}{ccc}
    \toprule
    Model & \texttt{Answerer} & Acc.(\%)\\
    \midrule
    \multirow{3}{*}{\ours{}} & BLIP-2 & 55.8\\
    & MiniGPT4 & 52.8 \\
    & LLaVA & 53.2\\
    \bottomrule
\end{tabular}
}
\end{center}
\caption{Ablation of choice of \texttt{Answerer}. }
\label{tab:answerer}
\end{table}

\noindent\textbf{\texttt{Answerer} Choice.} As we mentioned before, \texttt{Answerer} can be any VLM capable of answering visual questions. We mainly ablate three VLMs: BLIP-2, MiniGPT4, and LLaVA. The result is shown in Tab. \ref{tab:answerer}. Although MiniGPT4 and LLaVA can follow instructions and understand the longer contexts, they are worse than BLIP-2 when answering questions related to detailed visual facts. This also echoes the limitation mentioned in their papers about hallucinations and the lack of spatial/geometric  understanding. Note that the BLIP we use is BLIP2-FlanT5-XL, which in our experiments, gives a similar performance as BLIP2-FlanT5-XXL.

\begin{table}[t]
\begin{center}
\scalebox{0.9}{
\begin{tabular}{ccc}
    \toprule
    Model & Caption from & Acc.(\%)\\
    \midrule
    \multirow{3}{*}{\ours{}} & BLIP-2 & 55.8\\
    & MiniGPT4 & 48.4 \\
    & LLaVA & 51.2\\
    \bottomrule
\end{tabular}
}
\end{center}
\caption{Ablation of generated captions. }
\label{tab:caption}
\end{table}

\noindent\textbf{Image Captions.} To efficiently search for the best caption in our method, we fix the VLM as BLIP2-FlanT5-XL and go through pre-trained VLMs without fine-tuning on any caption dataset. From the experimental results shown in Tab. \ref{tab:caption}, we see BLIP2 exhibits the best performance. It is interesting to observe that both MiniGPT4 and LLaVA have shown impressive captioning ability in their papers/demos, but they fail in our framework when compared with BLIP2. We further go through many captions generated from MiniGPT4/LLaVA and BLIP2 and find that MiniGPT4/LLaVA tends to generate longer, more informative but less accurate (or more hallucination) captions. In contrast, BLIP-2 tends to generate short, less informative, but also less mistaken captions. So the hallucinations/mistakes in MiniGPT4/LLaVA-generated captions tend to mislead the \texttt{Questioner} and \texttt{Reasoner} to wrong answers.

% \textbf{Different Pre-trained VLMs.} Through our ablation studies on VCR, we find using BLIP2-FlanT5-XL in both image captioning and sub-questions answering can yield the best performance in zero-shot VCR. \rs{pending for more analysis}

\section{Conclusion}
In this work, we identify critical problems with existing VQA methods, especially the lack of addressing zero-shot reasoning tasks and the false assumption potentially forcing models to answer questions without sufficient information. To address these issues, we propose IdealGPT to utilize LLMs to construct a multiple-passes framework among a \texttt{Questioner}, a \texttt{Answerer}, and a \texttt{Reasoner}. This framework ensures better interpretability of VLMs' reasoning and robustness to re-correctifying predictions such as hallucination. Additionally, the generalizability of our framework can be illustrated by the modularity and our superior zero-shot performance. Our extensive experiments prove the effectiveness of each component in IdealGPT and verify that it can outperform the best existing GPT-4-like models by an absolute 10\% in VCR and 15\% in SNLI-VE.

\section*{Limitations}
Although we have witnessed the strong performance of the proposed method, we still have some limitations. Firstly, answering key sub-questions correctly plays a significant role in the success of our system. Therefore, our final results are largely bottlenecked by the performance of pre-trained VLMs. Additionally, we just employ the general image caption to give ChatGPT an idea of what the image looks like. However, dense captioning might be more informative and better help ChatGPT deal with the unseen image. Besides, the prompts are manually designed by us and it is difficult to find the optimal prompt in a specific situation. It will be better if the prompt can be generated automatically.

\section*{Ethics Statement}
ChatGPT is pre-trained on the colossal corpus which is likely to contain potential racial and gender bias. Therefore, if someone finds our work interesting and would like to use it in a specific environment, we strongly suggest the user check the potential bias before usage. In addition, it is hard to control the generation of LLMs like ChatGPT. We should be aware of the potential problems caused by hallucinations.

\section*{Acknowledgements}
This work is supported by DARPA MCS program under Cooperative Agreement N66001-19-2-4032.

\bibliography{anthology,custom}
\bibliographystyle{acl_natbib}

\appendix

% \section{Appendix}
\section{Data Statistics}
\label{app:statistics}

\begin{table}[h]\normalsize
\setlength\tabcolsep{8pt}
\renewcommand{\arraystretch}{1}
\centering
\begin{tabular}{lc|cc}
\toprule
\multicolumn{2}{c|}{VL Tasks} & Number of Samples \\ 
\midrule
 \multirow{4}{*}{SNLI-VE} & C & 1664 \\
 & N & 1672 \\
 & E & 1664 \\
 & All & 5000 \\
 \midrule
 VCR & - & 5000 \\
\bottomrule
\end{tabular}
\caption{\label{tab:statistics}
Statistics of SNLI-VE and VCR (C: Contradiction, N: Neural, E: Entailment, VL: Vision-Language)
}
\end{table}

We randomly select 5000 VCR and SNLI-VE samples from respective val and dev split. Zero-shot learning is conducted on these samples.

\section{Details of Visual Entailment Results}\label{app:vedetail}

\begin{table}[h]\small
\begin{center}
\renewcommand\tabcolsep{3pt}
\scalebox{1}{
\begin{tabular}{llcccc}
    \toprule
    & & C & N & E & All \\
    \midrule
    & Random Guess & - & - & - & 33.3 \\
    \midrule
    \multirow{3}{*}{\rotatebox{90}{\lb{Sup.}}}& \lb{EVE-Image} \citep{xie2019visual} & \lb{71.0} & \lb{70.6} & \lb{73.1} & \lb{71.6} \\
    & \lb{UNITER} \citep{chen2020uniter} & \lb{-} & \lb{-} & \lb{-} & \lb{79.4} \\
    & \lb{OFA} \citep{wang2022ofa} & \lb{-} & \lb{-} & \lb{-} & \lb{91.0} \\
    \midrule
    \multirow{4}{*}{\rotatebox{90}{ZS.}} & BLIP-2 \cite{li2023blip} & - & - & - & - \\ 
    & MiniGPT4 \cite{zhu2023minigpt} & 3.2 & 3.2 & \textbf{99.0} & 35.1 \\
    & LLaVA \cite{liu2023visual} & 12.0 & \textbf{31.3} & 77.6 & 40.3 \\
    & \textbf{\ours{}(ours)} & \textbf{83.4} & 25.9 & 56.7 & \textbf{55.3} \\
    \bottomrule
\end{tabular}
}
\end{center}
\caption{Accuracy of SNLI-VE (ZS: Zero-Shot, Sup: Supervised, C: Contradiction, N: Neutral, E: Entailment)}
\label{tab:vedetail}
\end{table}

From the results shown in Tab.\ref{tab:vedetail}, we can observe that our proposed method outperforms MiniGPT4 and LLaVA by a large margin. Moreover, MiniGPT4 exhibits near-chance level performance. When it is faced with different samples, it always replies \textit{entailment}, which indicates that it doesn't obtain a strong reasoning ability to process and understand the visual entailment task. However, \ours{} can achieve strong zero-shot results and even surpass the supervised EVE-Image model in \textit{contradiction} category.

\section{Details of Prompts Used}
\label{app:prompts}
\begin{figure*}[t]
        \centering
        \includegraphics[width=0.56\textheight]{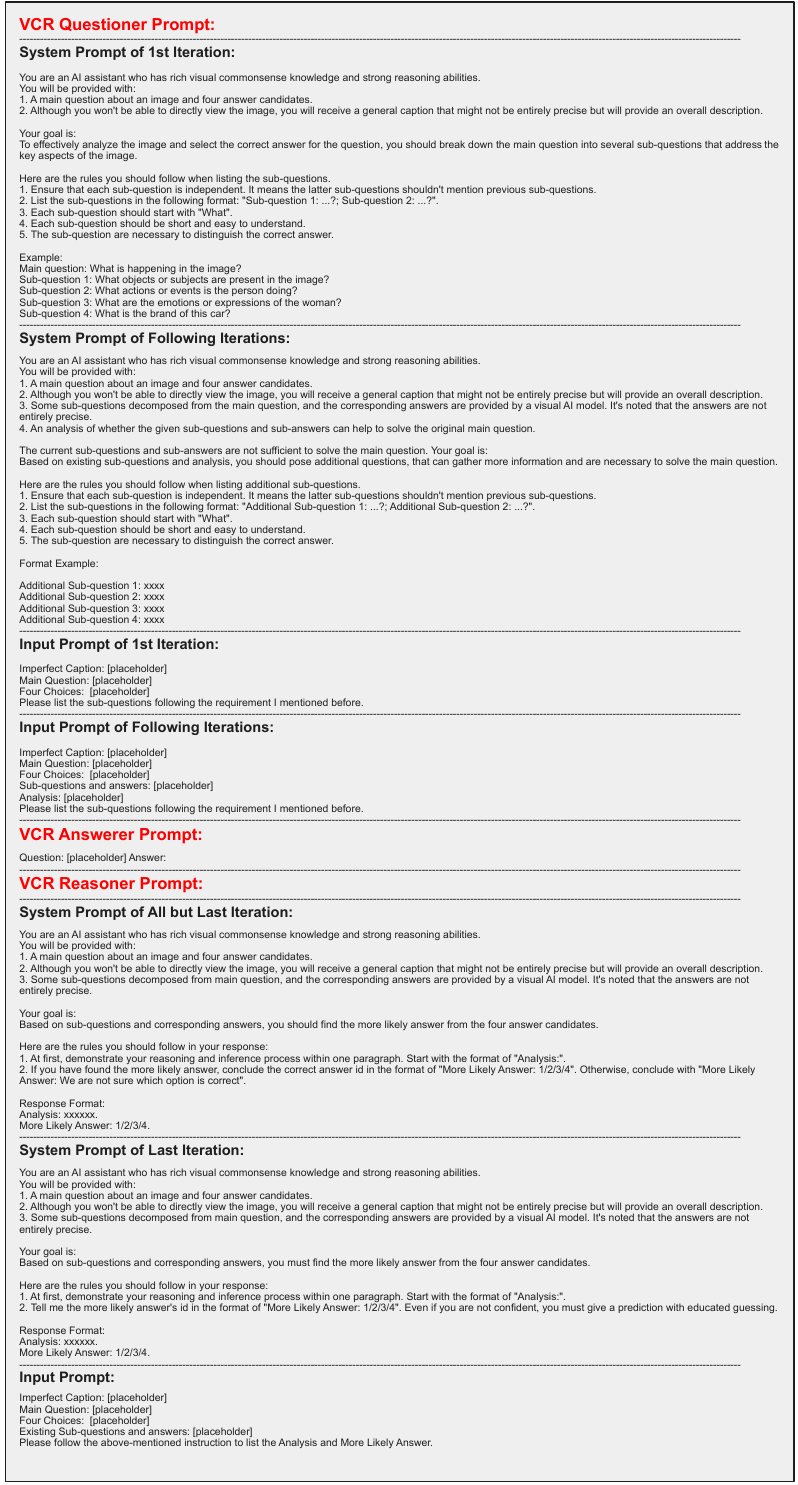}
        \caption{The prompts of \ours{} in VCR task.}
\label{fig:vcr_prompts}
\end{figure*}

\begin{figure*}[t]
        \centering
        \includegraphics[width=0.57\textheight]{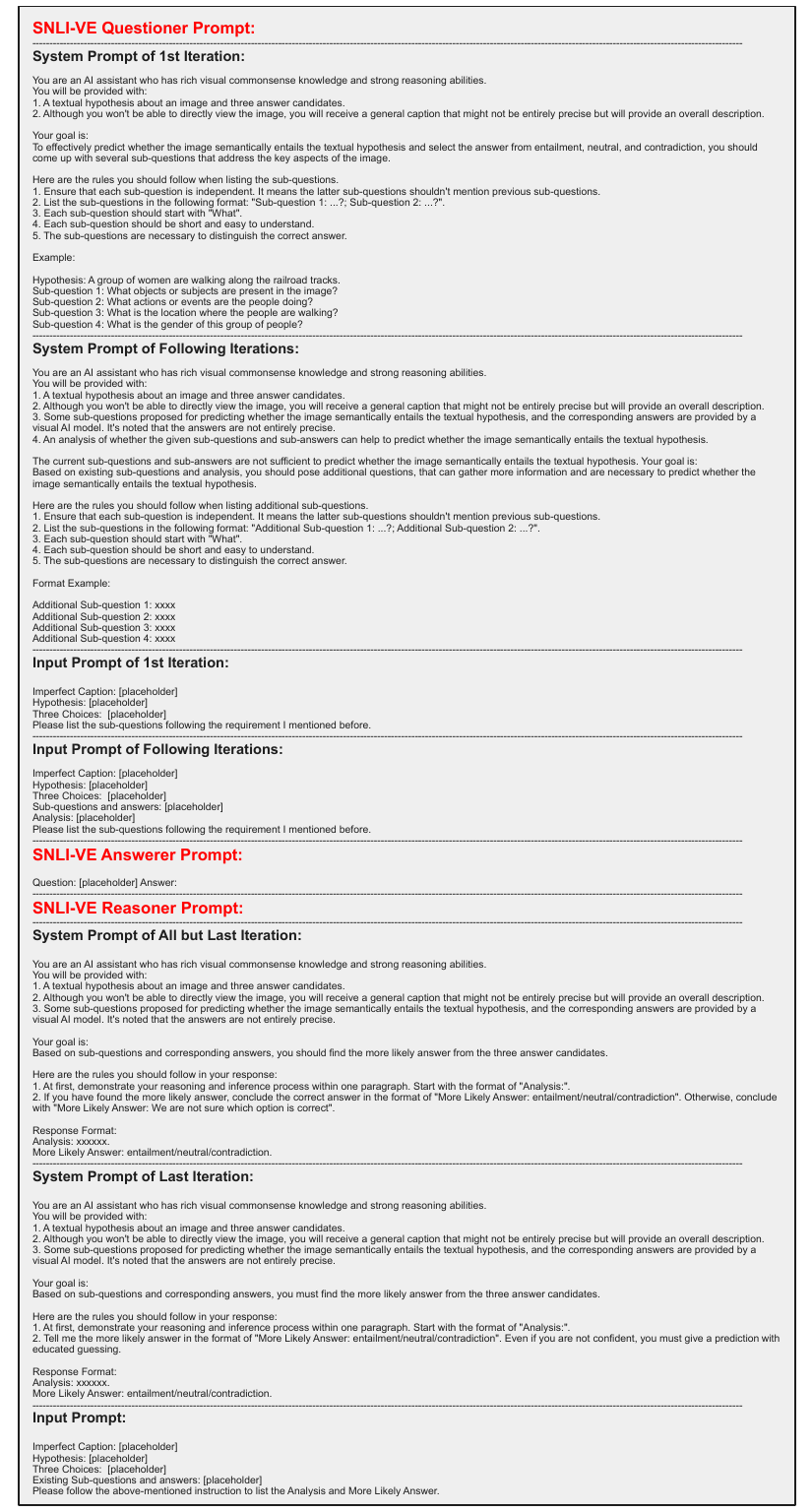}
        \caption{The prompts of \ours{} in SNLI-VE task.}
\label{fig:ve_prompts}
\end{figure*}

The prompts of \ours{} used in the VCR task are shown in Fig. \ref{fig:vcr_prompts}. As for SNLI-VE, the prompts are shown in Fig. \ref{fig:ve_prompts}. It's noted that the [placeholder] means we will replace it with the corresponding text of the instance, such as a main question, caption, and four choices. 

\section{VCR Pre-Processing}
\label{app:vcr_process}
We divide the image region from left to right into three bins and check if the mentioned object's center point belongs to which bin. If it's in the most left bin, it's renamed as ``person on the left''. Similarly,  ``person in the middle'' is in the middle bin, and ``person on the right'' is in the right bin. Since most QAs in VCR mentioned less than three persons, it can cover most cases.

\end{document}